\documentclass[lettersize,journal]{IEEEtran}

\usepackage{amsmath,amsfonts}
\usepackage{amssymb}    
\usepackage{algorithmic}

\usepackage{graphicx}
\usepackage[caption=false,font=footnotesize]{subfig} 

\usepackage{array}
\usepackage{booktabs}   
\usepackage{multirow}   
\usepackage[table]{xcolor} 

\usepackage{textcomp}
\usepackage{stfloats}
\usepackage{url}
\usepackage{verbatim}
\usepackage{cite}

\newcommand{\best}[1]{\textbf{#1}}
\newcommand{\second}[1]{\underline{#1}}
\newcommand{\xmark}{\textcolor{gray}{$\times$}}
\newcommand{\cmark}{\textcolor{black}{\checkmark}}
\definecolor{myblue}{RGB}{235, 245, 255}
\definecolor{goodRed}{RGB}{200, 0, 0}
\definecolor{badBlue}{RGB}{0, 100, 150}

\begin{document}

\title{Breaking the Resolution Barrier: Arbitrary-resolution Deep Image Steganography Framework}

\author{Xinjue Hu, Chi Wang, Xiang Zhang,~\IEEEmembership{Member,~IEEE,}, Boyu Wang, Zhenshan Tan, Zhangjie Fu,~\IEEEmembership{Member,~IEEE} 
\thanks{This work was supported in part by the National Natural Science Foundation of China under grant U22B2062, 62172232, by the Jiangsu Provincial Science and Technology Major Project (No. BG2024042), by the China Postdoctoral Science Foundation numbers 2023M741778. (Corresponding author: Zhangjie Fu).}
\thanks{Xinjue Hu, Chi Wang, Xiang Zhang, Boyu Wang, Zhenshan Tan and Zhangjie Fu are with the Engineering Research Center of Digital Forensics, Ministry of Education, Nanjing University of Information Science and Technology, Nanjing, 210044, China. (e-mail: xinjueh@126.com; 202412200715@nuist.edu.cn; zhangxiang@nuist.edu.cn; 202412200714@nuist.edu.cn; zstan@nuist.edu.cn; fzj@nuist.edu.cn).}
\thanks{Manuscript received April 19, 2021; revised August 16, 2021.}}

\markboth{Journal of \LaTeX\ Class Files,~Vol.~14, No.~8, August~2021}%
{Shell \MakeLowercase{\textit{et al.}}: A Sample Article Using IEEEtran.cls for IEEE Journals}

\IEEEpubid{0000--0000/00\$00.00~\copyright~2021 IEEE}

\maketitle

\begin{abstract}

Deep image steganography (DIS) has achieved significant results in capacity and invisibility. However, current paradigms enforce the secret image to maintain the same resolution as the cover image during hiding and revealing. This leads to two challenges: secret images with inconsistent resolutions must undergo resampling beforehand which results in detail loss during recovery, and the secret image cannot be recovered to its original resolution when the resolution value is unknown. To address these, we propose ARDIS, the first Arbitrary Resolution DIS framework, which shifts the paradigm from discrete mapping to reference-guided continuous signal reconstruction. Specifically, to minimize the detail loss caused by resolution mismatch, we first design a Frequency Decoupling Architecture in hiding stage. It disentangles the secret into a resolution-aligned global basis and a resolution-agnostic high-frequency latent to hide in a fixed-resolution cover. Second, for recovery, we propose a Latent-Guided Implicit Reconstructor to perform deterministic restoration. The recovered detail latent code modulates a continuous implicit function to accurately query and render high-frequency residuals onto the recovered global basis, ensuring faithful restoration of original details. Furthermore, to achieve blind recovery, we introduce an Implicit Resolution Coding strategy. By transforming discrete resolution values into dense feature maps and hiding them in the redundant space of the feature domain, the reconstructor can correctly decode the secret's resolution directly from the steganographic representation. Experimental results demonstrate that ARDIS significantly outperforms state-of-the-art methods in both invisibility and cross-resolution recovery fidelity.

\end{abstract}


\section{Introduction}
\IEEEPARstart{I}{mage} steganography aims to hide secret information within a cover image imperceptibly while ensuring the receiver can recover the secret information with high fidelity. Due to its covert nature, it has been widely applied in copyright protection \cite{editguard}, digital forensics \cite{omniguard}, military communications \cite{communication}, etc. Early steganographic methods typically relied on manually crafted embedding distortion functions \cite{pevny2010using,li2014investigation, holub2014universal, su2020image, 10323203}, which inherently suffered from limited adaptability and suboptimal performance. With the development of deep learning, techniques such as generative adversarial networks \cite{tang2017automatic, yang2019embedding,huang2023steganography, 10306313} and reinforcement learning \cite{tang2020automatic, mctsteg, reload,joint} have been utilized to enhance steganographic security. However, the payload capacity of these methods remains severely limited. To address the growing demand for high capacity information hiding, deep image steganography (DIS) has emerged as a promising paradigm, focusing on the concealment of entire, semantically meaningful secret images.

\begin{figure}[t]
  \centering
  \includegraphics[width=\columnwidth]{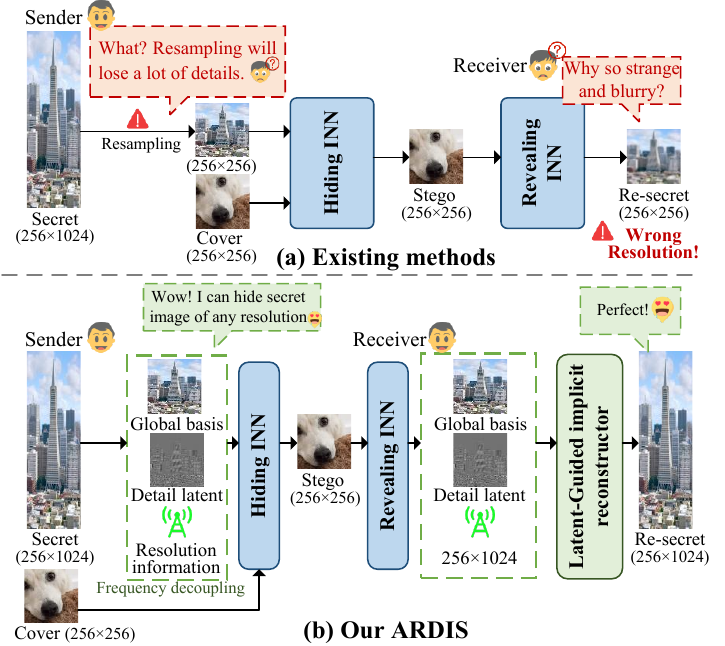}
  \caption{Overview of our ARDIS compared with existing methods. (a) Existing methods: The sender must manually resample the secret image to the same resolution as the cover image to hide it, and the recovered secret image will look strange due to the different resolution (e.g., different aspect ratio). (b) Our ARDIS enables arbitrary-resolution hiding and high-fidelity blind restoration.}
  \label{fig2}
\end{figure}
Existing DIS methods are mainly based on three architectures. Autoencoder-based methods \cite{baluja2017} rely on end-to-end encoder-decoder networks, which are prone to information loss due to repeated downsampling or nonlinear transformations. Methods based on invertible neural networks (INNs) \cite{ISN} model the hiding and revealing of the secret image as a pair of bijective mappings with shared parameters, which theoretically avoids information loss and thus enables near-lossless high-quality hiding. Recently, some works have introduced diffusion models \cite{diffusion} into steganography \cite{cRoss}, using prompts to guide the hiding and revealing process. However, their random sampling nature often limits pixel-level reconstruction accuracy.
\IEEEpubidadjcol
Despite their architectural differences, existing methods share a critical limitation: the \textbf{fixed-resolution constraint}. They formulate DIS as a discrete pixel-mapping problem, learning local mappings on spatially aligned pixels. This paradigm requires the secret and cover images to own the same dimension. This constraint presents two challenges in real-world applications, as shown in Figure \ref{fig2}. \textbf{First, Resolution Mismatch Dilemma}: When the secret image differs in resolution from the cover, especially when the resolution of the secret image is much larger than that of the cover image, lossy resampling of the secret image is necessary. This leads to the irreversible loss of high-frequency details. Simply applying post-hoc Super-Resolution methods(SR) is fundamentally inadequate. SR methods optimize statistical plausibility rather than instance fidelity, inevitably introducing generic hallucinations. This constitutes a fatal flaw for deep image steganography which demands precise recovery. \textbf{Second, Blind Recovery Incapacity}:  The receiver cannot obtain the resolution information of the secret image in advance, which means they can only recover a secret image with the same resolution as the cover image, but it is wrong. Relying on external transmission of metadata violates the principle of covert communication.

To bridge this gap, we propose ARDIS, the first Arbitrary-Resolution DIS framework. First, in order to solve the problem of detail loss caused by forced resampling under resolution mismatch at the sender side, our design is based on an intuitive idea: global low-frequency information is robust to resampling, but high-frequency details are highly susceptible. If they are processed as a whole, the fragile high-frequency details will inevitably be damaged due to resolution alignment constraints. Therefore, we propose frequency decoupling framework to decouple the secret image into a global visual basis and high-frequency detail latent information. The former characterizes global semantic cues and is aligned with the cover image’s resolution. The latter carries high-frequency residuals in a compressed, resolution-agnostic representation. This decoupling mechanism eliminates the impact of alignment constraints, ensuring that high-frequency details are not involved in lossy operations. Second, we find that the problem of lost detail in the recovered secret image caused by the limitations of the discrete pixel mapping paradigm: due to the limitations of fixed resolution, it must rely on interpolation operations when recovering secrets with wrong resolutions. To overcome this limitation, we shift the DIS paradigm from discrete pixel mapping to reference-guided continuous signal reconstruction, and design a latent-guided implicit reconstructor, which operates as a deterministically conditioned continuous neural field. We utilize the extracted high-frequency latent information as a conditional prior to modulate the implicit function, projecting the precise original high-frequency residuals onto the global basis to obtain a high-fidelity arbitrary-resolution secret image. Finally, to enable our reconstructor to perform blind recovery without violating steganography protocols, we introduce an implicit resolution encoding strategy. Simply hiding the resolution as a discrete scalar is extremely fragile, because the recovery process is not lossless, and small numerical fluctuations can lead to errors in resolution metadata. Therefore, we utilize the spatial redundancy of feature channels, broadcasting the resolution data into a dense feature map and embedding it, which provides a certain degree of fault tolerance. The receiver can adaptively infer the correct original resolution from the stego image, eliminating the dependence on external metadata.

Our contributions can be summarized as follows:
\begin{enumerate}
\item{To our knowledge, the proposed ARDIS is the first arbitrary-resolution DIS framework, effectively breaking the long-standing fixed-resolution constraint.}
\item{We design a frequency decoupling architecture for structure-detail separation of secret images at arbitrary resolutions. It encodes high-frequency textures that cannot be spatially aligned into a resolution-agnostic latent representation, thereby removing the dependence on cover-secret resolution alignment.}
\item{We propose a latent-guided implicit reconstructor that reformulates DIS recovery process as continuous image reconstruction based on a high-frequency detail latent, enabling high-fidelity texture recovery at arbitrary resolutions.}
\item{We introduce an implicit resolution coding strategy that leverages spatial redundancy in the feature domain to implicitly embed resolution priors, achieving geometry-consistent blind recovery when the secret resolution is unknown.}

\end{enumerate}

\section{Related Work}
\subsection{Autoencoder-based DIS methods}
Baluja \cite{baluja2017,baluja2019} first introduced deep learning into image steganography, modeling the task as an end-to-end encoding-decoding process. Subsequent researchers have improved steganographic performance by designing various network architectures. For instance, Wu et al. \cite{stegnet} added residual connections to the encoding network and used variance loss to reduce noise embedding in non-textured regions of the steganographic image, but color distortion was severe. Duan et al. \cite{Duan} used the U-Net \cite{unet} network as the steganographic network structure, and greatly improved the visual quality of the stego by utilizing its excellent feature extraction capabilities. Yu \cite{ABDH} proposed added an attention mechanism to the steganographic network to guide the secret information to be hidden in the complex textured regions of the cover image. Zhang et al. \cite{UDH} proposed a general DIS method that separates the secret image encoding process from the cover image. Ke et al. \cite{StegFormer} proposed the StegFormer framework based on autoencoders and Transformers \cite{transformer}, and achieved high visual quality image hiding by introducing channel-adaptive Transformer blocks and normalized training strategies. Liu et al. \cite{liu2025fearless} introduced dual-tree complex wavelet transform
(DTCWT) to enhance the richness of frequency representation and introduced the Mamba \cite{mamba} model to model long-distance dependencies and attention guidance, achieving high-quality image hiding. HGIS \cite{HGIS} enables generative steganography of multiple images. However, since the parameters of the hiding and revealing networks are not shared, it is difficult to achieve a perfect balance between the visual imperceptibility of the stego image and the recovery fidelity of the secret image.

\subsection{INNs-based DIS methods}
To address the aforementioned trade-off, inspired by the success of Invertible Neural Networks (INNs) \cite{Nice,realnvp} in fields such as image translation \cite{inn-trans}, researchers have attempted to reconstruct the steganographic task by leveraging the strict invertibility of INNs. \cite{ISN,HiNet,RIIS,DenseJIN,LiDiNet,stegmamba,SSHR,AIS}. For example, ISN \cite{ISN} first exploited this property to model the hiding and revealing processes as reciprocal mathematical transformations. HiNet \cite{HiNet} further enhanced the hiding capacity for high-frequency information by incorporating wavelet transforms. Xu et al. \cite{RIIS} took into account channel distortions occurring during the transmission of the stego image and proposed a robust steganography scheme. Li et al. \cite{LiDiNet} improved the information fusion mechanism to significantly reduce the number of model parameters while maintaining high steganographic capacity. Wang et al. \cite{SSHR} proposed a secure generative image hiding scheme that combines a reference image with an adaptive key, thereby enhancing the imperceptibility of the stego image and the security of the hidden secret image. Some researchers are considering further increasing the hiding capacity. For example, DeepMIH \cite{deepmih}, iSCMIS\cite{iSCMIS}, AIS \cite{AIS}, and StegFlow \cite{stegflow} have achieved hiding of multiple secret images. Although these methods achieve near-lossless hiding and recovery, they strictly rely on the spatial alignment of input and output dimensions, rendering them incapable of handling cross-scale steganography scenarios with mismatched resolutions.

\subsection{Diffusion-based DIS methods}
Recently, with the rapid development of AIGC, Diffusion Models \cite{ddpm} have been introduced into the steganography domain \cite{cRoss,diffstega,chen2025robust,jiang2026image} due to their powerful capability in modeling data distributions. CRoSS \cite{cRoss} leveraged the randomness in the generation of diffusion models to design a training-free steganography scheme. DiffStega \cite{diffstega} further added a preset password to avoid the risk of text prompts leaking information.  RGS-CM \cite{chen2025robust} achieves robust generative steganography through three reversible concatenated mappings. However, diffusion models are fundamentally probabilistic generation processes, and their reverse sampling process possesses inherent randomness. This results in uncontrollable detail hallucinations or deviations during the revealing process, making it difficult to achieve pixel-level precise restoration.

\begin{figure*}[t]
  \centering
  \includegraphics[width=\textwidth]{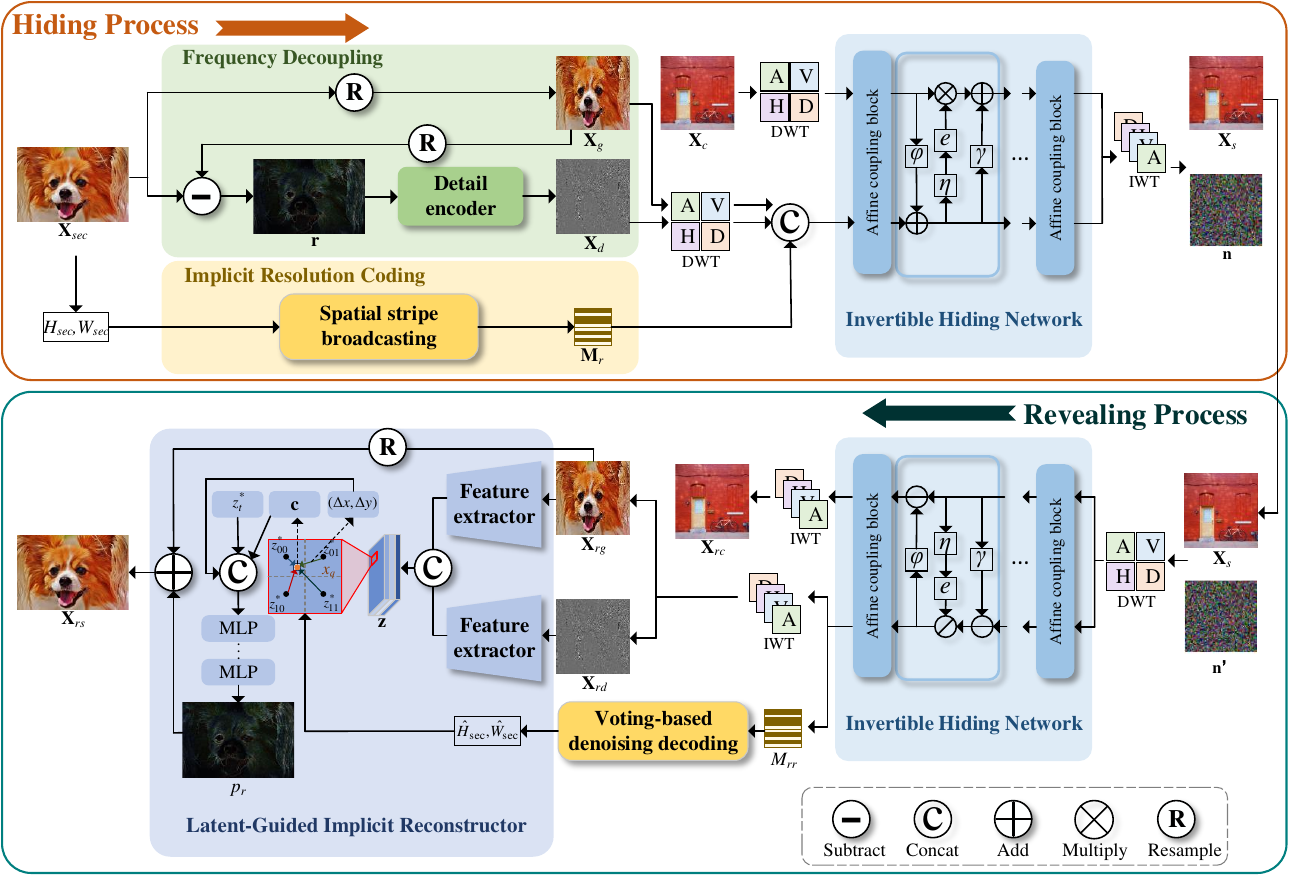}
  \caption{\textbf{Overview of the proposed ARDIS,} which supports hiding and revealing secret images at arbitrary resolutions. }
  \label{fig3}
\end{figure*}

\section{Proposed Method}

\subsection{Overview}
Our method consists of two stages: hiding and revealing. In the hiding process, shown in Figure \ref{fig3} (top), we first decouple the arbitrary-resolution secret image $\mathbf{X}_{sec}$ into a global visual basis $\mathbf{X}_g$, which matches the resolution of the cover $\mathbf{X}_c$, and a high-frequency detail latent code $\mathbf{X}_d$ via the frequency decoupling architecture (FDA). Simultaneously, the spatial dimensions value $(H_{sec}, W_{sec})$ are encoded into a feature map $\mathbf{M}_{r}$ via the implicit resolution coding strategy (IRC). $\mathbf{X}_c$, $\mathbf{X}_g$, and $\mathbf{X}_d$ are converted into frequency features using discrete wavelet transform (DWT), and then fed along with $\mathbf{M}_r$ into an invertible hiding network (IHN) consisting of a series of hidden blocks. The output of the last hidden block of the IHN is then go through an inverse wavelet transform (IWT) to generate the stego image $\mathbf{X}_s$ and the lost information $\mathbf{n}$. In the revealing process, shown in Figure \ref{fig3} (bottom), $\mathbf{X}_s$ and auxiliary variable $\mathbf{n'}$ are processed by DWT, inverse IHN, and IWT operations to obtain the recovered global visual basis $\mathbf{X}_{rg}$, detail latent code $\mathbf{X}_{rd}$, and resolution map $\mathbf{M}_{rr}$. $M_{rr}$ is decoded into the original dimensions via a resolution decoder. Subsequently, the recovered secret image $\mathbf{X}_{rs}$ is reconstructed by latent-guided implicit reconstructor (LGIR). Explicitly conditioned on the detail latent, the reconstructor performs continuous spatial queries to predict fine-grained residuals $p_r$, enabling high-fidelity secret image reconstruction at arbitrary target resolutions. FDA, LGIR, and IRC are detailed in Sections 3.2, 3.3, and 3.4, respectively.

\subsection{Frequency Decoupling Architecture}
The critical challenge of arbitrary resolution DIS lies in how to hide secret images at different resolutions into fixed-resolution cover images without losing information. We propose a Frequency Decoupling Architecture (FDA) to address it. The theoretical basis of FDA is that global semantic structure is spatially redundant, so the semantic topology of images remains largely intact even after resampling to the cover image’s resolution. Furthermore, the high-frequency residuals that determine fine textures are spatially sparse and extremely sensitive to resolution. Therefore, we utilize their sparsity to encode resolution-agnostic latent detail information. This strategy effectively bypasses strict pixel-level alignment constraints while preserving fine details.

Formally, Given a secret image $\mathbf{X}_{sec} \in \mathbb{R}^{H_{sec} \times W_{sec} \times 3}$ and a cover image $\mathbf{X}_c \in \mathbb{R}^{H_c \times W_c \times 3}$, we define an adaptive resampling operator $\mathcal{R}(\mathbf{X}, (H, W))$, which can adjust the input $\mathbf{X} \in \mathbb{R}^{H_{x} \times W_{x} \times 3}$ to the target size $(H, W)$ through bicubic interpolation. This operator can automatically perform downsampling (when $H_{x}>H$ or $W_{x}>W$) and upsampling (when $H_{x}<H$ or $W_{x}<W$) according to the actual situation.
First, we extract the global visual basis $\mathbf{X}_g$:

 \begin{align}
\mathbf{X}_{g} = \mathcal{R}(\mathbf{X}_{sec}, (H_{c}, W_{c})),
 \end{align}
where $\mathbf{X}_g$ is resolution-aligned with $\mathbf{X}_c$, preserving the basic global topological structure. It's worth noting that we choose explicit resampling over a general global extractor for this operation. This is because a learnable extractor introduces complex nonlinear global distortions, whereas resampling is a linear operation that strictly limits subsequently decoupled detail residuals to high frequencies and is spatially sparse. This is a prerequisite for subsequent detail encoding operations. 

Then, we calculate the precise detail loss $\mathbf{r}\in \mathbb{R}^{H_{sec} \times W_{sec} \times 3}$:
 \begin{align}
\mathbf{r}=\mathbf{X}_{sec}-\mathcal{R}(\mathbf{X}_g, (H_{sec}, W_{sec})).
 \end{align}

By explicitly resampling $\mathbf{X}_{g}$ back to the original resolution and calculating the residual, we can precisely capture the pixel-level detail loss during the alignment process. $\mathbf{r}$ is spatially sparse, mainly existing at edges and texture variations, thus it has high compressibility. Finally, we utilize a lightweight detail encoder $E_{detail}$ to encode $r$ into a compact high-frequency detail latent:

\begin{align}
\mathbf{X}_d = E_{detail}(r),
\end{align}
where $\mathbf{X}_d \in \mathbb{R}^{H_c \times W_c \times c_{lat}}$. Through adaptive spatial processing, $E_{detail}$ transforms the arbitrary-resolution $\mathbf{r}$ into the fixed resolution latent. Although $\mathbf{X}_d$ is spatially aligned with the cover for embedding, it is effectively used as a resolution-agnostic continuous prior for the subsequent reconstruction.

\subsection{Latent-Guided Implicit Reconstructor}
We propose the Latent Guided Implicit Reconstructor (LGIR), redefining the recovery process of the arbitrary resolution secret image as a reference guided deterministic continuous signal reconstruction. This ensures high fidelity recovery of the secret image. Specifically, we define a conditional decoding function $\phi_{\theta}$ to map continuous 2D coordinates $x \in \mathbb{R}^{2}$ to RGB values $s \in \mathbb{R}^{3}$ of the secret image. The mapping is formulated as follows:
\begin{equation}s = \phi_{\theta} \big( \mathcal{E}_{g}(X_{rg}(x)), \delta, c \ \big| \ \mathcal{E}_{d}(X_{rd}) \big)\end{equation}

where $c$ represents the cell decoding factors, $\delta$ denotes the relative coordinate, and $\mathcal{E}_{g}$ and $\mathcal{E}_{d}$ are feature encoders. The explicitly decoupled detail latent $\mathcal{E}_{d}(X_{rd})$ acts as a deterministic conditional mechanism rather than a generic spatial feature. Unlike standard implicit functions that rely solely on low resolution inputs to infer missing textures through statistical plausibility, our LGIR utilizes this detail latent as a hard boundary to perform deterministic spatial projection. This formulation ensures that the implicit function is strictly regularized to reconstruct authentic textures instead of generating generic hallucinations. 

To recover the original resolution secret image $X_{rs}$, we define the query coordinates $x_{q}$ at the target continuous scale. We denote $z_{t}^{*}$ as the local instance of the modulated latent code surrounding the query point, which intrinsically encapsulates both the global basis and the detail condition. The final RGB value at $x_{q}$ is deterministically reconstructed by querying the nearest surrounding $z_{t}^{*}$ as follows:
\begin{equation}X_{rs}(x_{q}) = \sum_{t} \omega_{t} \phi_{\theta}(z_{t}^{*}, x_{q}-v_{t}^{*}, c) + \mathcal{R}(X_{rg}, (H_{sec}, W_{sec}))(x_{q})\end{equation}

In this equation, $v_{t}^{*}$ is the spatial coordinate of the modulated feature $z_{t}^{*}$, and the relative coordinate is instantiated as $\delta = x_{q}-v_{t}^{*}$. The weight $\omega_{t}$ is calculated based on the area of the rectangle formed by the coordinates. This specific mathematical boundary explicitly compels the implicit function to synthesize missing textures guided by the authentic high frequency latent. This effectively guarantees faithful restoration across continuous scales while actively suppressing hallucinatory artifacts.

\subsection{Implicit Resolution Coding Strategy}
Another challenge arising from the misalignment of the secret image and the cover image's resolution is that the receiver needs to know the spatial size to reconstruct secret images correctly. However, transmitting this size metadata directly is not secure. To address this, we propose an Implicit Resolution Coding Strategy (IRC) to enable our reconstructor to has the ability to perform blind reconstruction. Instead of treating the resolution parameter as external metadata, we encode resolution attributes into a pseudo-visual feature map. This feature map is embedded along with other secret components, effectively transforming discrete numerical propagation into continuous feature regression. This allows the decoder to directly infer the original resolution from the stego image, ensuring image reconstruction without any metadata required.

During the hiding process, for the resolution $(H_{sec},W_{sec})$ of the secret image, we first quantize it into a L-bit binary sequence $\mathbf{b} \in \left \{0,1  \right \}^L $. To resist potential distortion from subsequent IHN convolution operations, we design a spatial stripe broadcasting mechanism to generate a resolution feature map $\mathbf{M}_{r} \in \mathbb{R}^{\frac{H_c}{2} \times\frac{W_c}{2}}$. This map is spatially aligned with the cover features after DWT and is partitioned into $L$ disjoint horizontal stripes, denoted as $\left\{\Omega_{k}\right\}_{k=1}^{L}$. For the $k$-th bit $b_k$ in $\mathbf{b}$, we map it to a specific stripe region $\Omega_k$ in the $\mathbf{M}_{r}$:
\begin{align} 
\mathbf{M}_{r}(x, y) = 2b_k - 1, \quad \forall (x, y) \in \Omega_k.
\end{align}
where the mapped values are normalized to $\left\{-1,1   \right\}$,which ensures the zero-mean property, thus aligning with the underlying spatial distribution of the subsequent IHN.

During the revealing process, since the recovered resolution map $\mathbf{M}_{rr}$ contain some distortion $\epsilon$, we design a voting-based denoising decoding mechanism:
\begin{equation}
\hat{b}_k = \mathbb{I}\left( \left(\frac{1}{|\Omega_k|} \sum_{(x,y)\in\Omega_k} \mathbf{M}_{rr}(x, y)\right) > 0 \right).
\end{equation}
where $ \mathbb{I}(\cdot)$is defined as:

\begin{equation}
\mathbb{I}(\mathcal{C}) =
\begin{cases}
1, & \text{if } \mathcal{C} \text{ is True} \\
0, & \text{otherwise}
\end{cases}.
\end{equation}

This mechanism performs collective voting and leverages the law of large numbers to cancel out randomly distributed distortion $\epsilon$, thereby enabling bit-by-bit accurate recovery of $\hat{b}_k$. This ensures zero-error recovery of the original resolution $(\hat{H}_{sec},\hat{W}_{sec}) $ of the secret image.
\begin{table*}[t]
\centering
\caption{Controlled resolution performance evaluation results. ``Blind Rec.'' signifies the ability to recover the secret image without explicit resolution metadata. \textbf{It is important to note that to calculate these visual metrics, we explicitly passed the resolution metadata to the contrastive methods that could not be blindly recovered. However, this is not permitted in real world scenarios.}}
\label{controll_table}

\setlength{\tabcolsep}{2.5pt} 
\renewcommand{\arraystretch}{1.1} 
\setlength{\aboverulesep}{0.1pt} 
\setlength{\belowrulesep}{0.1pt}
\resizebox{\textwidth}{!}{
\begin{tabular}{l c  | ccc ccc | ccc ccc | ccc ccc |c}
\toprule

\multirow{4}{*}{\textbf{Methods}} & 
\multirow{4}{*}{\shortstack{\textbf{Blind}\\\textbf{Rec.}}} & 
\multicolumn{18}{c|}{\textbf{DIV2K}}&
\multirow{4}{*}{\shortstack{\textbf{Avg.}\\\textbf{RRE}$\%$\\\\$\downarrow$}}\\
\cmidrule(lr){3-20}

& &  
\multicolumn{6}{c|}{\textbf{Res: $64 \times 64$}} & 
\multicolumn{6}{c|}{\textbf{Res: $128 \times 128$}} & 
\multicolumn{6}{c|}{\textbf{Res: $256 \times 256$}} \\
\cmidrule(lr){3-8} \cmidrule(lr){9-14} \cmidrule(lr){15-20}

& &  
\multicolumn{3}{c}{Stego} & \multicolumn{3}{c|}{Resecret} & 
\multicolumn{3}{c}{Stego} & \multicolumn{3}{c|}{Resecret} & 
\multicolumn{3}{c}{Stego} & \multicolumn{3}{c|}{Resecret} \\
\cmidrule(lr){3-5} \cmidrule(lr){6-8} \cmidrule(lr){9-11} \cmidrule(lr){12-14} 
\cmidrule(lr){15-17} \cmidrule(lr){18-20}

& &  
\scriptsize PSNR$\uparrow$ & \scriptsize SSIM$\uparrow$  & \scriptsize LPIPS$\downarrow$ & \scriptsize PSNR$\uparrow$ & \scriptsize SSIM$\uparrow$ & \scriptsize LPIPS$\downarrow$ &
\scriptsize PSNR$\uparrow$ & \scriptsize SSIM$\uparrow$ & \scriptsize LPIPS$\downarrow$ & \scriptsize PSNR$\uparrow$ & \scriptsize SSIM$\uparrow$ & \scriptsize LPIPS$\downarrow$ &
\scriptsize PSNR$\uparrow$ & \scriptsize SSIM$\uparrow$ & \scriptsize LPIPS$\downarrow$ & \scriptsize PSNR$\uparrow$ & \scriptsize SSIM$\uparrow$ &\scriptsize LPIPS$\downarrow$  \\
\midrule

ISN &  \xmark & 
\second{49.81} & \best{0.9971} & 0.0003& 41.71 & 0.9918 & 0.0068 & 
\second{49.25} & \second{0.9966} & 0.0004 & 36.22 & 0.9744 & 0.0355 & 
46.79 & 0.9938 & 0.0006 & 42.98 &0.9894 & 0.0018 & 
63.12\\

HiNet &  \xmark & 
45.99 & 0.9930 & 0.0002 & \second{42.16} & 0.9923 & 0.0066 & 
45.20 & 0.9916 & 0.0002 & \second{36.37} & \best{0.9755} & 0.0353 & 
41.69 & 0.9790 & 0.0007 & \second{46.16} & \second{0.9942} & \second{0.0011} & 
63.12 \\

StegFormer &  \xmark & 
47.91 & \second{0.9956} & 0.0003 & 39.44 & \second{0.9927} & 0.0097 & 
47.75 & 0.9954 & 0.0003 & 35.34 & 0.9745 & 0.0380 & 
\second{47.21} & \second{0.9948} & 0.0003 & 40.78 & 0.9924 & 0.0030 & 
63.12 \\

AIS &  \xmark & 
46.53 & 0.9664 & \second{0.0001} & 41.59 & 0.9886 & \second{0.0061} & 
45.81 & 0.9637 & 0.0001 & 36.14 & 0.9728 & \second{0.0313} & 
44.57 & 0.9574 & \second{0.0002} &33.22 & 0.8875 & 0.0149 & 
63.12 \\

\rowcolor{myblue}
\textbf{ARDIS} & \cmark & 
\best{50.30} & \best{0.9971} & \best{0.0000} & \best{42.51} & \best{0.9930} & \best{0.0048} & 
\best{49.94} & \best{0.9968} & \best{0.0000} & \best{36.46} & \second{0.9746} & \best{0.0300} & 
\best{48.09} & \best{0.9951} & \best{0.0001} & \best{48.39} & \best{0.9966} & \best{0.0001} & 
\best{0.00} \\

\midrule

\multirow{3}{*}{\textbf{Methods}} & 
\multirow{3}{*}{\shortstack{\textbf{Blind}\\\textbf{Rec.}}} & 
\multicolumn{6}{c|}{\textbf{Res: $512 \times 512$}} & 
\multicolumn{6}{c|}{\textbf{Res: $720 \times 720$}} & 
\multicolumn{6}{c|}{\textbf{Res: $1024 \times 1024$}} &
\multirow{3}{*}{\shortstack{\textbf{Avg.}\\\textbf{RRE}$\%$\\\\$\downarrow$}}\\
\cmidrule(lr){3-8} \cmidrule(lr){9-14} \cmidrule(lr){15-20}

& &  
\multicolumn{3}{c}{Stego} & \multicolumn{3}{c|}{Resecret} & 
\multicolumn{3}{c}{Stego} & \multicolumn{3}{c|}{Resecret} & 
\multicolumn{3}{c}{Stego} & \multicolumn{3}{c|}{Resecret} \\
\cmidrule(lr){3-5} \cmidrule(lr){6-8} \cmidrule(lr){9-11} \cmidrule(lr){12-14} 
\cmidrule(lr){15-17} \cmidrule(lr){18-20}

& &  
\scriptsize PSNR$\uparrow$ & \scriptsize SSIM$\uparrow$  & \scriptsize LPIPS$\downarrow$ & \scriptsize PSNR$\uparrow$ & \scriptsize SSIM$\uparrow$ & \scriptsize LPIPS$\downarrow$ &
\scriptsize PSNR$\uparrow$ & \scriptsize SSIM$\uparrow$ & \scriptsize LPIPS$\downarrow$ & \scriptsize PSNR$\uparrow$ & \scriptsize SSIM$\uparrow$ & \scriptsize LPIPS$\downarrow$ &
\scriptsize PSNR$\uparrow$ & \scriptsize SSIM$\uparrow$ & \scriptsize LPIPS$\downarrow$ & \scriptsize PSNR$\uparrow$ & \scriptsize SSIM$\uparrow$ &\scriptsize LPIPS$\downarrow$  \\
\midrule

ISN &  \xmark & 
45.81 & 0.9922 & 0.0008 & \second{27.88} & 0.8379 & 0.2380 & 
44.64 & 0.9896 & 0.0010 & \second{26.05} & 0.7728 & 0.3425 & 
44.27 & 0.9885 & 0.0012 & \second{24.80} & 0.7076 & 0.4507 &
63.12\\

HiNet &  \xmark & 
39.80 & 0.9680 & 0.0016 & \second{27.88} & \second{0.8400} & 0.2371 & 
37.73 & 0.9508 & 0.0030 & \second{26.05} & \second{0.7739} & 0.3421 & 
37.14 & 0.9440 & 0.0042 & 24.79 & 0.7083 & 0.4578 &
63.12 \\

StegFormer &  \xmark & 
\second{47.02} & \best{0.9946} & 0.0003 & 27.74 & 0.8361 & \second{0.2339} & 
\best{46.68} & \best{0.9941} & \second{0.0004} & 25.98 & 0.7725 & \second{0.3324} & 
\best{46.56} & \best{0.9939} & \second{0.0004} & 24.74 & \second{0.7085} & \second{0.4391} &
63.12 \\

AIS &  \xmark & 
44.27 & 0.9566 & \second{0.0002} & 26.33 & 0.7516 & 0.3122 &
43.83 & 0.9535 & \best{0.0002} & 24.67 & 0.6629 & 0.4202 &
43.63 & 0.9522 & \best{0.0002} & 23.63 & 0.6048 & 0.5318 &
63.12 \\

\rowcolor{myblue}
\textbf{ARDIS} & \cmark & 
\best{47.23} & \second{0.9940} & \best{0.0001} & \best{30.93} & \best{0.8917} & \best{0.1027} & 
\second{46.28} & \second{0.9923} & \best{0.0002} & \best{28.08} & \best{0.8151} & \best{0.1775} & 
\second{45.81} & \second{0.9913} & \best{0.0002} & \best{26.63} & \best{0.7516} & \best{0.2607} &
\best{0.00} \\

\midrule
\midrule

\multirow{4}{*}{\textbf{Methods}} & 
\multirow{4}{*}{\shortstack{\textbf{Blind}\\\textbf{Rec.}}} & 
\multicolumn{18}{c|}{\textbf{Flickr2K}}&
\multirow{4}{*}{\shortstack{\textbf{Avg.}\\\textbf{RRE}$\%$\\\\$\downarrow$}}\\
\cmidrule(lr){3-20}

& &  
\multicolumn{6}{c|}{\textbf{Res: $64 \times 64$}} & 
\multicolumn{6}{c|}{\textbf{Res: $128 \times 128$}} & 
\multicolumn{6}{c|}{\textbf{Res: $256 \times 256$}} \\
\cmidrule(lr){3-8} \cmidrule(lr){9-14} \cmidrule(lr){15-20}

& &  
\multicolumn{3}{c}{Stego} & \multicolumn{3}{c|}{Resecret} & 
\multicolumn{3}{c}{Stego} & \multicolumn{3}{c|}{Resecret} & 
\multicolumn{3}{c}{Stego} & \multicolumn{3}{c|}{Resecret} \\
\cmidrule(lr){3-5} \cmidrule(lr){6-8} \cmidrule(lr){9-11} \cmidrule(lr){12-14} 
\cmidrule(lr){15-17} \cmidrule(lr){18-20}

& &  
\scriptsize PSNR$\uparrow$ & \scriptsize SSIM$\uparrow$  & \scriptsize LPIPS$\downarrow$ & \scriptsize PSNR$\uparrow$ & \scriptsize SSIM$\uparrow$ & \scriptsize LPIPS$\downarrow$ &
\scriptsize PSNR$\uparrow$ & \scriptsize SSIM$\uparrow$ & \scriptsize LPIPS$\downarrow$ & \scriptsize PSNR$\uparrow$ & \scriptsize SSIM$\uparrow$ & \scriptsize LPIPS$\downarrow$ &
\scriptsize PSNR$\uparrow$ & \scriptsize SSIM$\uparrow$ & \scriptsize LPIPS$\downarrow$ & \scriptsize PSNR$\uparrow$ & \scriptsize SSIM$\uparrow$ &\scriptsize LPIPS$\downarrow$  \\
\midrule

ISN &  \xmark & 
\second{49.38} & \best{0.9970} & 0.0003 & 42.51 & 0.9905 & 0.0071 & 
\second{48.88} & \second{0.9966} & 0.0004& 37.45 & 0.9745 & 0.0372 & 
\second{46.60} & \second{0.9937} & 0.0007 & 42.54 & 0.9858 & 0.0028 & 
63.12\\

HiNet &  \xmark & 
45.29 & 0.9926 & 0.0003 & \second{43.56} & \second{0.9927} & 0.0069 & 
44.72 & 0.9770 & 0.0003 & \second{37.92} & \second{0.9770} & 0.0369 & 
41.63 & 0.9781  & 0.0009 & \second{45.86} & \second{0.9928} & \second{0.0008} & 
63.12 \\

StegFormer &  \xmark & 
47.11 & 0.9942 & 0.0005 & 40.18 & 0.9920 & \best{0.0010} & 
47.01 & 0.9941 & 0.0005 & 36.25 & 0.9751 & 0.0396 & 
46.53 & 0.9933 & 0.0005 & 40.90 & 0.9906 & 0.0034 & 
63.12 \\

AIS &  \xmark & 
45.90 & 0.9514 & \second{0.0002} & 42.84 & 0.9766 & 0.0058 & 
45.42 & 0.9499 & \second{0.0002} & 37.78 & 0.9689 & \second{0.0311} & 
44.35 & 0.9436 & \second{0.0002} & 34.45 & 0.8811 & 0.0145 & 
63.12 \\

\rowcolor{myblue}
\textbf{ARDIS} & \cmark & 
\best{49.61} & \second{0.9969} & \best{0.0000} & \best{44.19} & \best{0.9937} & \second{0.0050} & 
\best{49.32} & \best{0.9967} & \best{0.0000} & \best{38.16} & \best{0.9765} & \best{0.0300} & 
\best{47.68} & \best{0.9949} & \best{0.0001} & \best{48.20} & \best{0.9960} & \best{0.0001} & 
\best{0.00} \\

\midrule

\multirow{3}{*}{\textbf{Methods}} & 
\multirow{3}{*}{\shortstack{\textbf{Blind}\\\textbf{Rec.}}} & 
\multicolumn{6}{c|}{\textbf{Res: $512 \times 512$}} & 
\multicolumn{6}{c|}{\textbf{Res: $720 \times 720$}} & 
\multicolumn{6}{c|}{\textbf{Res: $1024 \times 1024$}} &
\multirow{3}{*}{\shortstack{\textbf{Avg.}\\\textbf{RRE}$\%$\\\\$\downarrow$}}\\
\cmidrule(lr){3-8} \cmidrule(lr){9-14} \cmidrule(lr){15-20}

& &  
\multicolumn{3}{c}{Stego} & \multicolumn{3}{c|}{Resecret} & 
\multicolumn{3}{c}{Stego} & \multicolumn{3}{c|}{Resecret} & 
\multicolumn{3}{c}{Stego} & \multicolumn{3}{c|}{Resecret} \\
\cmidrule(lr){3-5} \cmidrule(lr){6-8} \cmidrule(lr){9-11} \cmidrule(lr){12-14} 
\cmidrule(lr){15-17} \cmidrule(lr){18-20}

& &  
\scriptsize PSNR$\uparrow$ & \scriptsize SSIM$\uparrow$  & \scriptsize LPIPS$\downarrow$ & \scriptsize PSNR$\uparrow$ & \scriptsize SSIM$\uparrow$ & \scriptsize LPIPS$\downarrow$ &
\scriptsize PSNR$\uparrow$ & \scriptsize SSIM$\uparrow$ & \scriptsize LPIPS$\downarrow$ & \scriptsize PSNR$\uparrow$ & \scriptsize SSIM$\uparrow$ & \scriptsize LPIPS$\downarrow$ &
\scriptsize PSNR$\uparrow$ & \scriptsize SSIM$\uparrow$ & \scriptsize LPIPS$\downarrow$ & \scriptsize PSNR$\uparrow$ & \scriptsize SSIM$\uparrow$ &\scriptsize LPIPS$\downarrow$  \\
\midrule

ISN &  \xmark & 
45.66 & 0.9921 & 0.0008 & 28.66 & 0.8501 & 0.2238 & 
44.57 & 0.9896 & 0.0011 & 26.73 & 0.7909 & 0.3142 & 
44.19 & 0.9887 & 0.0012 & 25.15 & 0.7218 & 0.4225 & 
63.12\\

HiNet &  \xmark & 
39.91 & 0.9680 & 0.0018 & \second{28.72} & \second{0.8532} & 0.2233 & 
37.98 & 0.9518 & 0.0035 & \second{26.76} & \second{0.7930} & 0.3147 & 
37.37 & 0.9469 & 0.0043 & \second{25.17} & \second{0.7229} & 0.4303 &
63.12 \\

StegFormer &  \xmark & 
\second{46.40} & \second{0.9931} & 0.0005 & 28.44 & 0.8484 & \second{0.2214} & 
\second{46.12} &\best{ 0.9926} & \second{0.0006} & 26.60 & 0.7903 & \second{0.3066} & 
\best{46.03} & \best{0.9924} & 0.0006 & 25.07 & 0.7220 & \second{0.4145}&
63.12 \\

AIS &  \xmark & 
44.18 & 0.9426 & \second{0.0002} & 27.02 & 0.7620 & 0.2944& 
43.82 & 0.9403 & \best{0.0002} & 25.23 & 0.6821 & 0.3932& 
43.67 & 0.9391 & \best{0.0002} & 23.95 & 0.6202 & 0.5147&
63.12 \\

\rowcolor{myblue}
\textbf{ARDIS} & \cmark & 
\best{46.93} & \best{0.9939} & \best{0.0001} & \best{31.88} & \best{0.9022} & \best{0.0929}& 
\best{46.13} & \second{0.9925} & \best{0.0002} & \best{28.73} & \best{0.8304} & \best{0.1611}& 
\second{45.73} & \second{0.9918} & \second{0.0003} & \best{26.97} & \best{0.7636} & \best{0.2478}&
\best{0.00} \\
\bottomrule
\end{tabular}
}
\end{table*}

\section{Experimental Results}

\subsection{Datasets and Implementation Details}

\subsubsection{Implementation Details}We implement the framework using PyTorch on an NVIDIA Tesla V100 GPU. The training process employs the Adam optimizer with an initial learning rate of $10^{-4}$ and a Cosine Annealing decay schedule over 800 epochs. A channel-wise learnable clamp in IHN is initialized at 2.0 to stabilize high-frequency embedding. The MLP decoder in LGIR consists of 4 layers with a width of 256 using ReLU activations.

\subsubsection{Evaluation Protocols}To establish an arbitrary resolution DIS evaluation benchmark, we constructed a multi-protocol evaluation benchmark to evaluate the invisibility of the stego image and the fidelity of the recovered secret image. We conduct extensive experiments across four public datasets: DIV2K \cite{div2k}, Flickr2K \cite{flickr2k}, COCO \cite{coco}, and Stego260 \cite{cRoss}.

\textbf{Protocol-1: Controlled Resolution Performance Evaluation.} Leveraging the high resolution datasets DIV2K and Flickr2K, this protocol establishes a strictly controlled evaluation environment free from interpolation artifacts. To facilitate a comprehensive assessment across scenes of arbitrary resolution, we fix the resolution of the cover images at 256$\times$256, while the secret images are dynamically cropped to encompass two distinct scenarios: ``small-in-large" and ``large-in-small". Specifically, the resolutions utilized include 64$\times$64, 128$\times$128, 256$\times$256, 512$\times$512, 720$\times$720, and 1024$\times$1024.

\textbf{Protocol-2: Generalization Evaluation in Complex Scenes.} This evaluates the model's adaptability to realistic, complex textured scenes and its ability to recover detail fidelity of the secret image across diverse image aspect ratios. This protocol utilizes the COCO dataset. Unlike lossless high-definition datasets such as DIV2K, the COCO dataset includes real-world image compression, lighting variations, and other factors. In the experiment, the cover images were still set to a resolution of 256$\times$256, while the secret images retained their original resolutions.

\textbf{Protocol-3: Domain-Specific Evaluation.} Diffusion-based DIS methods achieve exceptional visual quality without requiring training. To comprehensively evaluate the differences between our method and these state-of-the-art paradigms, a specialized evaluation protocol is essential. Standard image datasets lack the carefully crafted text prompts. Therefore, we employ the Stego260 dataset. This dataset is derived from public datasets \cite{dataset1,dataset2} and the Google search engine and contains carefully crafted prompts from CRoSS.

\subsubsection{Benchmarks} To validate the superiority of our ARDIS, we compared it with state-of-the-art DIS methods, including ISN \cite{ISN}, HiNet \cite{HiNet}, CRoSS \cite{cRoss}, StegFormer \cite{StegFormer}, DiffStega \cite{diffstega}, and AIS \cite{AIS}.

\subsubsection{Evaluation strategy} Since existing methods cannot perform blind recovery, we implement an \textbf{asymmetric evaluation strategy: the contrasting methods are evaluated under settings that explicitly provide the true resolution, while our method is evaluated under a blind recovery setting}. We used PSNR, SSIM, and LPIPS to evaluate hiding and recovery performance. Furthermore, we defined relative resolution error (RRE) specifically to evaluate our blind recovery capability:
\begin{align}
RRE=\frac{1}{2}\left(\frac{\left|\hat{H}-H_{g t}\right|}{H_{g t}}+\frac{\left|\hat{W}-W_{g t}\right|}{W_{g t}}\right) \times 100 \%.
\end{align}
where $\hat{H}$ and $\hat{W}$ are predicted resolution, $H_{g t}$ and $W_{g t}$ are ground-truth resolution.

\begin{figure*}[t]
\centering
\includegraphics[width = \textwidth]{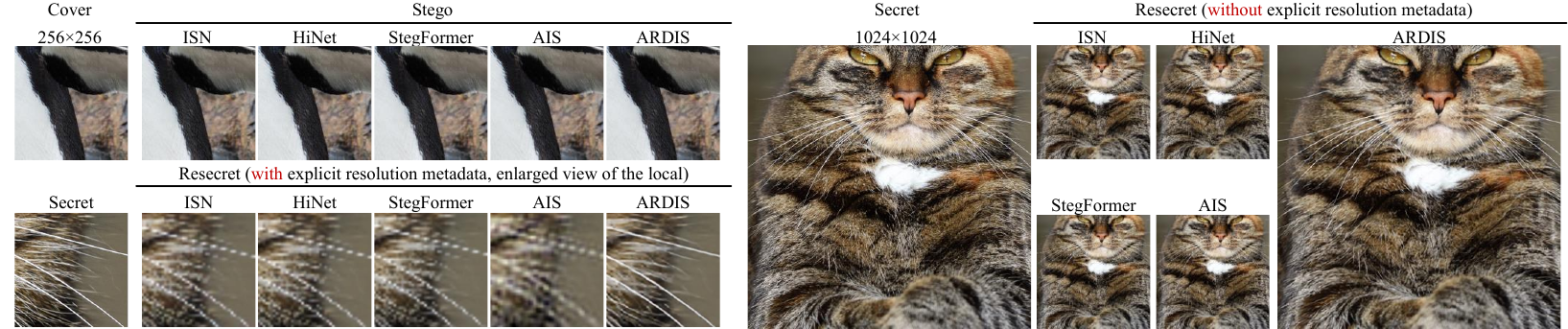}
\caption{Controlled resolution performance evaluation results: Visual comparisons of our ARDIS with leading deep image steganography methods for stego and recovered secret images in arbitrary-resolution hiding scenarios. The recovery of secret images presents two scenarios. (1) Without explicitly transmitting resolution metadata, no comparison methods can recover the secret image at the original resolution. (2) Explicitly transmitting resolution metadata, the comparison methods relies on resampling for recovery, which results in the loss of many details.}
\label{control_fig}
\end{figure*}

\begin{figure*}[t]
\centering
\includegraphics[width = \textwidth]{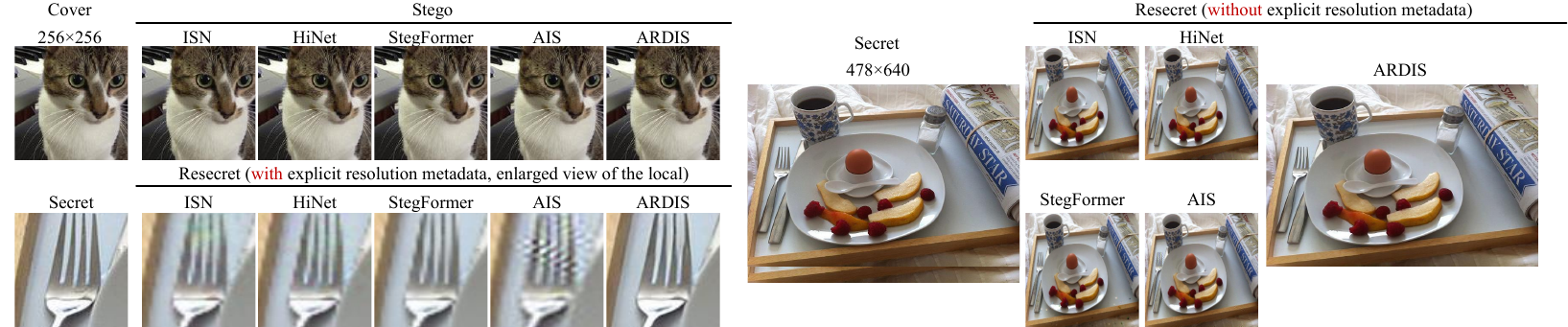}
\caption{Generalization evaluation results in Complex Scenes.}
\label{gene_fig}
\end{figure*}


\begin{figure}
\centering
\includegraphics{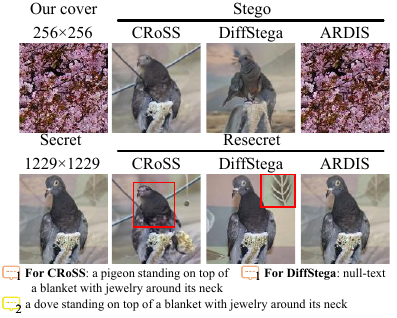}
\caption{The visual comparison of ARDIS and the diffusion-based DIS method on the Stego260 dataset. The diffusion-based DIS method requires no cover and simply follows the prompts to generate the corresponding stego image. The settings for prompts 1 and 2 completely follow the design of the CRoSS and Diffstega methods.}
\label{domain_2}
\end{figure}

\subsection{Experimental Results}

\subsubsection{Protocol 1: Controlled Resolution Performance Evaluation}
Table \ref{controll_table} and Figure \ref{control_fig} present the quantitative and qualitative results across diverse resolution conditions. Since existing benchmark methods naturally ensure high stego image quality, the primary performance differentiator is the recovery fidelity of the secret image. In scenarios where the secret resolution is smaller than the cover, ARDIS demonstrates exceptional stability. Specifically, when recovering a $64 \times 64$ secret, ARDIS outperforms the second best method by 0.35 dB in PSNR and 0.0003 in SSIM.

The advantages of ARDIS are particularly evident in extreme ``large-in-small” scenarios. When the secret image resolution reaches 1024$\times$1024, the payload is 16 times the spatial capacity of the cover image. Despite this immense information density, ARDIS still performs exceptionally well. On the DIV2K dataset, its PSNR value is 1.83 dB higher than that of the second-best method, its SSIM value is 0.044 higher, and its LPIPS value is 0.1784 lower.  Crucially, ARDIS is the unique solution capable of blind recovery with a 0$\%$ RRE, while other methods can only recover secret images at the same resolution as the cover image, resulting in an RRE as high as 63.12$\%$. The visualization results in Figure \ref{control_fig} further demonstrate the advantages of ARDIS. As can be seen, even when the comparison methods are manually provided with true resolution metadata to aid in reconstruction, high-frequency details are lost. For example, the cat’s whiskers appear severely blurred and jagged. In contrast, our ARDIS method preserves fine details. Furthermore, when the true resolution metadata is removed, the comparison methods can only produce severely downsampled images of no practical value, whereas ARDIS remains capable of accurately reconstructing any native resolution.

\begin{table}[h]
\centering
\caption{Generalization evaluation results on the COCO dataset.}
\label{general_table}

\renewcommand{\arraystretch}{1.1} 
\setlength{\aboverulesep}{0.1pt} 
\setlength{\belowrulesep}{0.1pt}
\resizebox{\columnwidth}{!}{
\begin{tabular}{l c  | ccc ccc |c }
\toprule

\multirow{3}{*}{\textbf{Methods}} & 
\multirow{3}{*}{\shortstack{\textbf{Blind}\\\textbf{Rec.}}} & 
\multicolumn{6}{c|}{\textbf{COCO}} &
\multirow{3}{*}{\shortstack{\textbf{Avg.}\\\textbf{RRE}$\%$\\\\$\downarrow$}}\\
\cmidrule(lr){3-8}

& & 
\multicolumn{3}{c}{Stego} & \multicolumn{3}{c|}{Resecret} \\
\cmidrule(lr){3-5} \cmidrule(lr){6-8}

& &
\scriptsize PSNR$\uparrow$ & \scriptsize SSIM$\uparrow$ & \scriptsize LPIPS$\downarrow$ & \scriptsize PSNR$\uparrow$ & \scriptsize SSIM$\uparrow$ & \scriptsize LPIPS$\downarrow$\\
\midrule

ISN &  \xmark & 
42.76 & 0.9873 & 0.0008 & 26.65 &0.7954 &0.2549 &49.36\\

HiNet &  \xmark & 
37.22 & 0.9556 & 0.0026 & \second{26.67} & \second{0.7975}  &\second{0.2540}&49.36\\

StegFormer &  \xmark & 
41.99 & \best{0.9869} & 0.0008 & 26.51 & 0.7846 &0.2549 &49.36\\

AIS & \xmark & 
\second{42.03} & 0.9254 & 0.0004 & 25.42 & 0.7022 &0.4054&49.36\\

\rowcolor{myblue}
\textbf{ARDIS} &\cmark & 
\best{42.65} & \best{0.9869} & \best{0.0002}& \best{28.70} & \best{0.8415} & \best{0.1115} &\best{0.00}\\

\bottomrule
\end{tabular}
}
\end{table}
\subsubsection{Protocol-2: Generalization Evaluation in Complex Scenes}Table \ref{general_table} and Figure \ref{gene_fig} demonstrate the model’s generalization capabilities on real-world data. As shown in Table \ref{general_table}, ARDIS exhibits a significant advantage during the revealing process. Specifically, the PSNR value for the recovered secret image is 2.03 dB higher than the second-best result, its SSIM value has improved by 0.044, and its LPIPS has been significantly reduced by 0.1425. In the visualization results shown in Figure \ref{gene_fig}, we selected a secret image with dimensions of 478$\times$640. Observing the local magnified view, even when provided with the true resolution metadata, the other comparison methods introduce severe artifacts and edge blurring. In contrast, our ARDIS perfectly restores the sharp edges and the true luster of the metal fork. After removing the resolution metadata, the other comparison methods can only compress the originally rectangular secret image into a square of the same resolution as the cover image, resulting in proportional distortion.

\begin{table}[h]
\centering
\caption{Domain-specific evaluation results.}
\label{domain-specific_table}

\renewcommand{\arraystretch}{1.1} 
\setlength{\aboverulesep}{0.1pt} 
\setlength{\belowrulesep}{0.1pt}
\resizebox{\columnwidth}{!}{
\begin{tabular}{l c  | ccc ccc |c}
\toprule

\multirow{3}{*}{\textbf{Methods}} & 
\multirow{3}{*}{\shortstack{\textbf{Blind}\\\textbf{Rec.}}} & 
\multicolumn{6}{c|}{\textbf{stego260}} &
\multirow{3}{*}{\shortstack{\textbf{Avg.}\\\textbf{RRE}$\%$\\\\$\downarrow$}}\\
\cmidrule(lr){3-8}

& &
\multicolumn{3}{c}{Stego} & \multicolumn{3}{c|}{Resecret} \\
\cmidrule(lr){3-5} \cmidrule(lr){6-8}

& & 
\scriptsize PSNR$\uparrow$ & \scriptsize SSIM$\uparrow$ & \scriptsize LPIPS$\downarrow$ & \scriptsize PSNR$\uparrow$ & \scriptsize SSIM$\uparrow$ & \scriptsize LPIPS$\downarrow$\\
\midrule

ISN & \xmark & 
\second{46.62} & \second{0.9939} & 0.0006 & 30.87 &0.8792 &0.2153&62.83\\

HiNet &  \xmark & 
41.05 & 0.9776 & 0.0011 & \second{30.98} & \second{0.8828}  &0.2172&62.83\\

CRoSS &  \xmark & 
-- & -- & -- & 22.58 & 0.7308  &0.2567&36.70\\

StegFormer & \xmark & 
46.52 & 0.9930 & 0.0005 & 30.66 & 0.8758 &0.2169 &62.83\\

DiffStega & \xmark & 
-- & -- & --  & 24.67 & 0.7621 &\second{0.2110} &36.70\\

AIS & \xmark & 
44.36 & 0.9456 & \second{0.0004} & 29.48 & 0.8175 &0.3009&62.83\\

\rowcolor{myblue}
\textbf{ARDIS} &  \cmark & 
\best{48.01} & \best{0.9954} & \best{0.0001}& \best{32.89} & \best{0.9018} & \best{0.1091} &\best{0.00}\\

\bottomrule
\end{tabular}
}
\end{table}

\subsubsection{Protocol-3: Domain-Specific Evaluation}Table \ref{domain-specific_table} presents the quantitative evaluation results. During the revealing process, ARDIS demonstrated superior performance across a range of secret image resolutions from 225$\times$225 to 3452$\times$3452. The PSNR value was 1.91 dB higher than that of the second best method and 8.22 dB higher than that of the best diffusion-based DIS method. The LPIPS was 0.1019 lower than that of the second best method. Furthermore, ARDIS remains the exclusive model to achieve a 0\% RRE. 

Figure \ref{domain_2} reveals a critical limitation of diffusion based methods, specifically the issue of semantic hallucinations. During the revealing process, while the images generated by these methods may appear visually realistic, they actually contain textures completely unrelated to the original secret image. For example, DiffStega fabricates a false leaf background in the pigeon image. In contrast, ARDIS generates high-quality stego images while ensuring pixel-level consistency between the recovered results and the original secret image.


\subsubsection{Security analysis} 
\begin{figure}[h]
\centering
\includegraphics[width = 0.70\columnwidth]{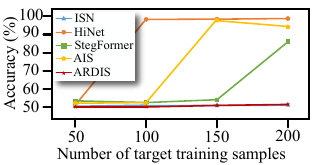}
\caption{Steganalysis accuracy by SRNet. The fact that the curve remains close to 50$\%$ despite the increasing number of leaked samples demonstrates the high security of the method.}
\label{fig_5}
\end{figure}

The anti-steganalysis ability is a important metric for evaluating the security of DIS methods. We assess the security of each method by exploring the number of leaked samples an attacker needs to effectively detect stego images\cite{weng}. A detection accuracy closer to 50$\%$ indicates higher security. We train and test SRNet \cite{SRNet} using cover / stego image pairs generated on the COCO dataset by our method or other comparison methods. As shown in Figure \ref{fig_5}, our method maintains superior security even with an increase in leaked samples.

\begin{table}[t]
\centering
\caption{Ablation experiments on the FDA module evaluated on the COCO dataset. We compared the full FDA (Variant \#4) with three other baseline models.}
\label{abl_fda}

\setlength{\tabcolsep}{2.5pt} 
\renewcommand{\arraystretch}{1} 
\setlength{\aboverulesep}{0.1pt} 
\setlength{\belowrulesep}{0.1pt}
\resizebox{\linewidth}{!}{

\begin{tabular}{l c | ccc | ccc}
\toprule

\multirow{2}{*}{\textbf{Variant}} & 
\multirow{2}{*}{\textbf{Methods}} & 
\multicolumn{3}{c|}{\textbf{Stego}} & \multicolumn{3}{c}{\textbf{Resecret}}  \\
\cmidrule(lr){3-5} \cmidrule(lr){6-8} 

& &  
\scriptsize PSNR$\uparrow$ & \scriptsize SSIM$\uparrow$  & \scriptsize LPIPS$\downarrow$ & \scriptsize PSNR$\uparrow$ & \scriptsize SSIM$\uparrow$ & \scriptsize LPIPS$\downarrow$ \\
\midrule

\#1 &  No FDA & 
40.48 & 0.9797 & 0.0004 & \second{28.27} & \second{0.8299} & \second{0.1278} \\

\#2 &  Laplacian Operator & 
\second{42.38} & \second{0.9865} & \best{0.0002} & 27.41 & 0.8128 & 0.1744 \\

\#3 &  Adaptive Max Pool & 
41.87 & 0.9854 & \best{0.0002} & 27.25 & 0.8079 & 0.1786 \\

\#4 &  FDA & 
\best{42.65} & \best{0.9869} & \best{0.0002} & \best{28.70} & \best{0.8415} & \best{0.1115} \\

\bottomrule
\end{tabular}
}
\end{table}

\subsection{Ablation Study}

In this section, we conduct comprehensive ablation experiments to validate the effectiveness of each ARDIS component. Specifically, we evaluate the designs of FDA, LGIR, and IRC using Protocol-2 on real world images.
\subsubsection{Effectiveness of FDA} 

To validate the effectiveness of the FDA module, we set up three other variants. Variant \#1 remove FDA module, \#2 replaces the detail encoder in FDA with a Laplacian operator, and \#3 replaces the detail encoder with a adaptive max pooling. Table \ref{abl_fda} presents the ablation results.
Removing the FDA (\#1) degrades stego image quality by 2.17 dB and recovered secret image quality by 0.43 dB due to the unmitigated loss of high frequency details. \#2 and \#3. perform even worse during the revealing process, illustrating the negative impact of erroneous detail priors on secret image recovery. Specifically, the Laplacian operator in Variant \#2 is a statically designed high-pass filter. It cannot adaptively determine which high-frequency features should be preserved based on the actual requirements of the reconstruction, resulting in a biased prior on the details provided to the LGIR. Although the adaptive max pooling in variant \#3 preserves the most significant details in the residuals, it disrupts spatial topological relationships and feature continuity. When these incomplete, discrete features are input into the LGIR module as conditional variables, the continuous mapping function cannot establish an accurate correspondence between coordinates and pixel values. Our FDA module, on the one hand, preserves details while eliminating the effects of alignment constraints. On the other hand, the adaptive detail encoder maps the decoupled high-frequency residuals to a continuity prior that is better suited to the LGIR module, thereby ensuring the fidelity of the recovered secret image.

\begin{figure}
\centering
\includegraphics[width = \columnwidth]{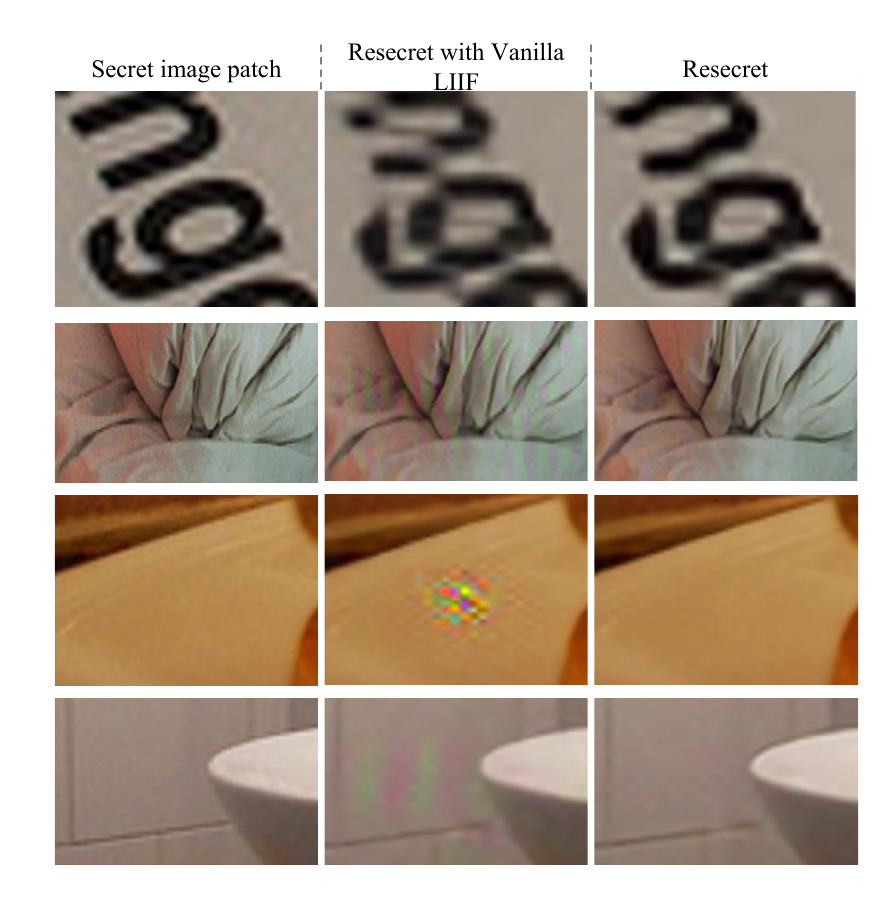}
\caption{Impact of latent guidance on detail reconstruction. With vanilla LIIF, the reconstruction fails to resolve high-frequency textures, leading to over-smoothed edges and chromatic aberrations.}
\label{fig_8}
\end{figure}

\subsubsection{Effectiveness of LGIR} 

To validate the effectiveness of the LGIR module, we set up two additional variants. Variant \#1 remove LGIR module (denote ``no LGIR"), \#2 replace the LGIR module with vanilla LIIF. As shown in Table \ref{abl_lgir}, when the LGIR module is completely removed (\#1), the high-frequency details extracted during the revealing process cannot be accurately aligned to the global visual basis, resulting in blurred details. While the introduction of Vanilla LIIF (\#2) improves the quality of the recovered secret image to some extent, it can only rely on global features to blindly guess missing textures due to the lack of precise high-frequency detail guidance. As shown in Figure \ref{fig_8}, it leads to color distortion or blurred details. In contrast, our LGIR module not onpy overcomes the limitations of discrete pixel mapping by utilizing continuous signal reconstruction but also employs deterministic detail guidance to restore authentic textures. This fully ensures the fidelity of the recovered secret image.

\begin{table}[t]
\centering
\caption{Ablation experiments on the LGIR module evaluated on the COCO dataset. We compared the full LGIR (Variant \#3) with two other baseline models.}
\label{abl_lgir}

\setlength{\tabcolsep}{2pt} 
\renewcommand{\arraystretch}{1} 
\setlength{\aboverulesep}{0.1pt} 
\setlength{\belowrulesep}{0.1pt}
\resizebox{\linewidth}{!}{

\begin{tabular}{l c | ccc | ccc}
\toprule

\multirow{2}{*}{\textbf{Variant}} & 
\multirow{2}{*}{\textbf{Methods}} & 
\multicolumn{3}{c|}{\textbf{Stego}} & \multicolumn{3}{c}{\textbf{Resecret}}  \\
\cmidrule(lr){3-5} \cmidrule(lr){6-8} 

& &  
\scriptsize PSNR$\uparrow$ & \scriptsize SSIM$\uparrow$  & \scriptsize LPIPS$\downarrow$ & \scriptsize PSNR$\uparrow$ & \scriptsize SSIM$\uparrow$ & \scriptsize LPIPS$\downarrow$ \\
\midrule

\#1 &  No LGIR & 
\second{42.08} & \second{0.9859} & \best{0.0002} & 27.12 & 0.8055 & \second{0.1605} \\

\#2 &  Vanilla LIIF & 
41.79 & 0.9846 & \second{0.0003} & \second{27.25} & \second{0.8071} & 0.1794 \\

\#3 &  LGIR & 
\best{42.65} & \best{0.9869} & \best{0.0002} & \best{28.70} & \best{0.8415} & \best{0.1115} \\

\bottomrule
\end{tabular}
}
\end{table}

\subsubsection{Effectiveness of IRC} 
\begin{figure}
\centering
\includegraphics[width = \columnwidth]{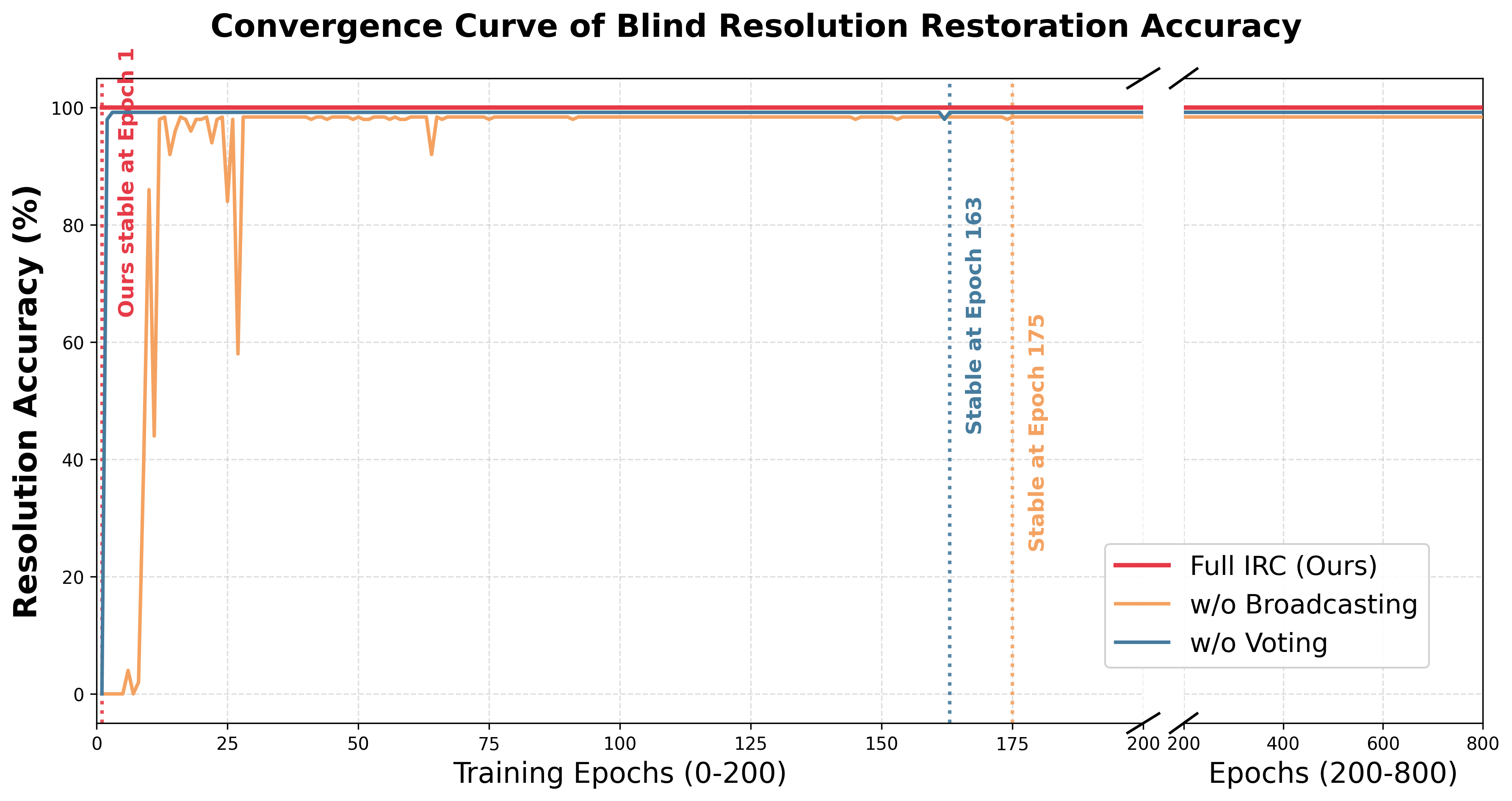}
\caption{The impact of different mechanisms in the IRC module on the accuracy of resolution recovery. Whether it is the absence of a broadcasting mechanism or a voting mechanism, the model requires a significant amount of time to recover the secret image resolution with 100\% accuracy.}
\label{blind_recovery}
\end{figure}

To validate the effectiveness of the IRC module, we set up two additional variants. Variant \#1: remove the broadcasting mechanism from IRC (denoted as ``no broadcasting”), \#2 remove the voting mechanism from IRC (denoted as ``no voting”). Figure \ref{blind_recovery} shows the accuracy of resolution data recovery for different variants during training. As can be seen, although all three variants ultimately recover the resolution correctly, both removing the broadcasting mechanism and removing the voting mechanism require a longer number of training epochs to stably output correct resolution values. This reflects that, in the absence of effective redundancy and fault-tolerance mechanisms, the model must satisfy strict continuous numerical regression constraints when fitting the resolution metadata. Due to the deep entanglement of features, this leads to a conflict in co-optimization between the resolution recovery task and the main image hiding and revealing tasks. As shown in Table \ref{abl_irc}, this conflict results in a significant decline in the visual quality of both the stego and resecret images for variants \#1 and \#2.  In contrast, our IRC achieves a stable 100\% accuracy rate as early as the first training round, allowing the model to focus more on optimizing the stego and recovered secret images, ultimately achieving optimal visual quality.

\begin{table}[t]
\centering
\caption{Ablation experiments on the IRC module evaluated on the COCO dataset. We compared the full IRC (Variant \#3) with two other baseline models.}
\label{abl_irc}

\setlength{\tabcolsep}{2pt} 
\renewcommand{\arraystretch}{1} 
\setlength{\aboverulesep}{0.1pt} 
\setlength{\belowrulesep}{0.1pt}
\resizebox{\linewidth}{!}{

\begin{tabular}{l c | ccc | ccc}
\toprule

\multirow{2}{*}{\textbf{Variant}} & 
\multirow{2}{*}{\textbf{Methods}} & 
\multicolumn{3}{c|}{\textbf{Stego}} & \multicolumn{3}{c}{\textbf{Resecret}}  \\
\cmidrule(lr){3-5} \cmidrule(lr){6-8} 

& &  
\scriptsize PSNR$\uparrow$ & \scriptsize SSIM$\uparrow$  & \scriptsize LPIPS$\downarrow$ & \scriptsize PSNR$\uparrow$ & \scriptsize SSIM$\uparrow$ & \scriptsize LPIPS$\downarrow$ \\
\midrule

\#1 &  No broadcasting & 
\second{41.57} & \second{0.9841} & \best{0.0002} & 27.30 & 0.8106 & 0.1724 \\

\#2 &  No voting & 
41.29 & 0.9833 & \second{0.0003} & \second{27.31} & \second{0.8112} & \second{0.1699} \\

\#3 &  IRC & 
\best{42.65} & \best{0.9869} & \best{0.0002} & \best{28.70} & \best{0.8415} & \best{0.1115} \\

\bottomrule
\end{tabular}
}
\end{table}

\subsubsection{Effectiveness of each module}We evaluate the effectiveness of each proposed module. We set up three variants: \#1 remove FDA and LGIR and keep only backbone, \#2 remove LGIR, \#3 remove FDA. We retain the IRC module across all tests to ensure the network can access the target resolution natively. The experimental results are shown in Table \ref{abl}.  Removing the FDA (\#3) forces standard resampling before hiding and leads irreversible high frequency information loss. The reconstructor then produces noticeable artifacts due to a complete lack of deterministic guidance. Furthermore, utilizing FDA without LGIR (\#2) actually decreases performance compared to the pure backbone (\#1). This indicates that standard resampling operators cannot effectively integrate the decoupled high-frequency latent variables. In contrast, introducing both FDA and LGIR (\#4) significantly improves the quality of the reconstructed secret image, highlighting the superiority of our design.

\begin{table}[t]
\centering
\caption{Ablation experiments of different components evaluated on the COCO dataset.}
\label{abl}

\setlength{\tabcolsep}{2pt} 
\renewcommand{\arraystretch}{1} 
\setlength{\aboverulesep}{0.1pt} 
\setlength{\belowrulesep}{0.1pt}
\resizebox{\linewidth}{!}{

\begin{tabular}{l ccc | ccc | ccc}
\toprule

\multirow{2}{*}{\textbf{Variant}} & 
\multirow{2}{*}{\textbf{FDA}} & 
\multirow{2}{*}{\textbf{LGIR}} & 
\multirow{2}{*}{\textbf{IRC}} & 
\multicolumn{3}{c|}{\textbf{Stego}} & \multicolumn{3}{c}{\textbf{Resecret}}  \\
\cmidrule(lr){5-7} \cmidrule(lr){8-10} 

& & & &  
\scriptsize PSNR$\uparrow$ & \scriptsize SSIM$\uparrow$  & \scriptsize LPIPS$\downarrow$ & \scriptsize PSNR$\uparrow$ & \scriptsize SSIM$\uparrow$ & \scriptsize LPIPS$\downarrow$ \\
\midrule

\#1 &  \xmark & \xmark & \cmark & 
\second{42.35} & \second{0.9863} & \second{0.0003} & 27.18 & 0.8070 & 0.1654 \\

\#2 &  \cmark & \xmark & \cmark & 
42.08 & 0.9859 & \best{0.0002} & 27.12 & 0.8055 & 0.1605 \\

\#3 &  \xmark & \cmark & \cmark & 
40.48 & 0.9797 & 0.0004 & \second{28.27} & \second{0.8299} & \second{0.1278} \\

\#4 &  \cmark & \cmark & \cmark & 
\best{42.65} & \best{0.9869} & \best{0.0002} & \best{28.70} & \best{0.8415} & \best{0.1115} \\
\bottomrule
\end{tabular}
}
\end{table}

\section{Conclusion}
In this paper, we propose \textbf{ARDIS}, the first framework to enable arbitrary-resolution deep image steganography. This work breaks the long-standing fixed-resolution constraint inherent in current DIS paradigms. Through the frequency decoupling architecture, ARDIS achieves the decoupled embedding of global structure and fine-grained detail during the hiding process, guaranteeing the effective preservation of high-frequency information. Meanwhile, via the latent-guided implicit reconstructor, it ensures faithful and determinative detail restoration in the revealing process. Furthermore, robust blind recovery is realized through the implicit resolution coding strategy. Experimental results demonstrate that ARDIS significantly outperforms state-of-the-art methods in both invisibility and cross-resolution recovery fidelity.

\bibliographystyle{IEEEtran}
\bibliography{reference.bib}

\vfill

\end{document}